\documentclass{article} 
\usepackage{iclr2027_conference,times}

\usepackage{amsmath,amsfonts,bm}

\def\eqref#1{equation~\ref{#1}}

\def\1{\bm{1}}

\DeclareMathAlphabet{\mathsfit}{\encodingdefault}{\sfdefault}{m}{sl}
\SetMathAlphabet{\mathsfit}{bold}{\encodingdefault}{\sfdefault}{bx}{n}

\usepackage{hyperref}
\usepackage{url}
\usepackage{graphicx}
\usepackage{amsmath,amssymb}
\usepackage{booktabs}
\usepackage{multirow}
\usepackage{xcolor}
\usepackage{colortbl}
\usepackage{cleveref}
\usepackage{algorithm}
\usepackage{algpseudocode}
\usepackage{wrapfig}
\usepackage{needspace}

\title{World SLAM Model: Joint World Modeling for SLAM and Navigation}

\author{
  \textbf{Minghui Qin$^{1,3*}$,
  Yijun Yuan$^{2*\dagger}$,
  Weicheng Zheng$^{2}$,
  Kenan Li$^{2}$,
  Weibang Wang$^{2}$,} \\
  \textbf{Chang Sun$^{2}$, 
  Junhao Huang$^{2}$, 
  Anmin Liu$^{2}$, 
  Yicheng Yao$^{2}$,
  Hang Zhao$^{1,2\dagger}$}\\
  $^1$Shanghai Qi Zhi Institute \\
  $^2$IIIS, Tsinghua University\\
  $^3$Shanghai Jiao Tong University\\
  \color{blue}{\tt\small{\url{https://tsinghua-mars-lab.github.io/WorldSLAMModel}}}\\
mhqin1018@gmail.com,{\{yuanyj, hangzhao\}@mail.tsinghua.edu.cn}\\
}
\iclrfinalcopy 

\begin{document}

\maketitle

\begin{abstract}
We introduce \textbf{World SLAM Model (WSM)}, a unified framework that brings
the SLAM paradigm directly into downstream navigation.
Rather than treating SLAM merely as an upstream module that provides poses,
maps or tokens, WSM adopts its core mechanisms, including incremental state updates with persistent memory and backend refinement of accumulated errors, to maintain a consistent world state during interaction.
Given the current observation and a navigation goal, WSM predicts future visual states and jointly estimates their camera motion and dense geometry, grounding visual prediction in an evolving spatial world state.
This spatial state is continuously updated as new observations arrive and provides the basis for action generation and closed-loop navigation.
WSM is trained end-to-end with a joint navigation--SLAM objective, enabling downstream navigation to benefit directly from SLAM-style state maintenance
and refinement while preserving accurate geometric estimation. 
Experiments demonstrate improved navigation performance together with strong SLAM accuracy, highlighting the potential of SLAM as an intrinsic mechanism for long-horizon world modeling and embodied interaction.
\end{abstract}

\section{Introduction}
\label{sec:introduction}

SLAM has long served as a fundamental spatial module for robotics~\citep{davison2007monoslam}.
Given a stream of observations, a SLAM system estimates camera motion and scene structure, producing poses and maps that are subsequently consumed by downstream tasks such as navigation and control~\citep{weiss2011monocular,cadena2016past}.
In a modular navigation pipeline, the SLAM module supplies poses and maps to downstream planning and control, while continuously maintaining and refining these estimates~\citep{chaplot2020learning}.

To reduce the complexity of this modular pipeline, recent navigation methods increasingly adopt end-to-end models that directly map visual observations to trajectories or actions~\citep{cheng2024navila,zhang2024navid,shah2023vint}. 
While this removes an explicit SLAM module from the navigation pipeline, it does not eliminate the dependence on spatial information traditionally provided by SLAM~\citep{yin2025unigoal}. StreamVLN~\citep{wei2025streamvln} and Dynam3D~\citep{wang2026dynam3d} still take depth as input, while NavDP~\citep{cai2025navdp} requires online robot poses or goal locations from external ground-truth localization. 
More recent approaches obtain such information from pretrained 3D reconstruction models~\citep{peng2025logoplanner,zeng2025janusvln}, but still treat geometry as an upstream estimate fed to the navigation module.

However, directly using SLAM outputs still leaves state estimation and navigation separately optimized, preventing navigation objectives from shaping how the world state is learned and maintained.
We argue that navigation should instead benefit from \emph{SLAM itself}, including its mechanisms for maintaining memory and correcting accumulated errors.
Integrating SLAM into the navigation model benefits both training and inference.
During training, geometric and navigation objectives jointly shape the shared world-state representation and its maintenance mechanisms.
During inference, shared streaming memory and error correction continuously update the state used for planning and action prediction.

World models provide a natural framework for realizing this idea.
A world model must continuously process observations, retain information from past interactions, predict future evolution, and update its state as new observations arrive.
These closely match the core mechanisms of SLAM, making SLAM a suitable computational structure for long-horizon world modeling rather than merely an external source of geometric estimates.

Based on this observation, we introduce the
\textbf{World SLAM Model (WSM)}, which brings the SLAM mechanism directly into downstream navigation.
Instead of following the conventional
\emph{SLAM $\rightarrow$ navigation} pipeline, WSM incorporates spatial estimation, memory, prediction, and action into a shared closed-loop model.
In this way, navigation benefits not only from what SLAM estimates, but also from how SLAM continuously maintains and corrects the world state.

Specifically, the SLAM expert first estimates depth and camera poses from the current observation. Conditioned on the goal, the generation expert then dreams future observations from which the same SLAM expert, acting as an inverse dynamics model, infers the future world states and recovers the camera trajectory. After the trajectory is executed, new observations are fed back into the model, forming a continuous prediction--interaction--update loop. A persistent KV cache enables incremental processing across interaction steps, while the SLAM expert maintains spatial consistency as observations accumulate.

We evaluate WSM on both closed-loop navigation and SLAM. On the InternVLA-N1~\citep{internvla-n1} start--goal navigation benchmark, WSM relies only on RGB observations and its own geometric estimates, yet achieves higher success rates than all evaluated baselines. On held-out InternVLN-N1 trajectories, we further evaluate its SLAM capability to verify that the model retains accurate spatial estimation while serving the downstream task.

\begin{figure}[t]
    \centering
    \includegraphics[width=\columnwidth]{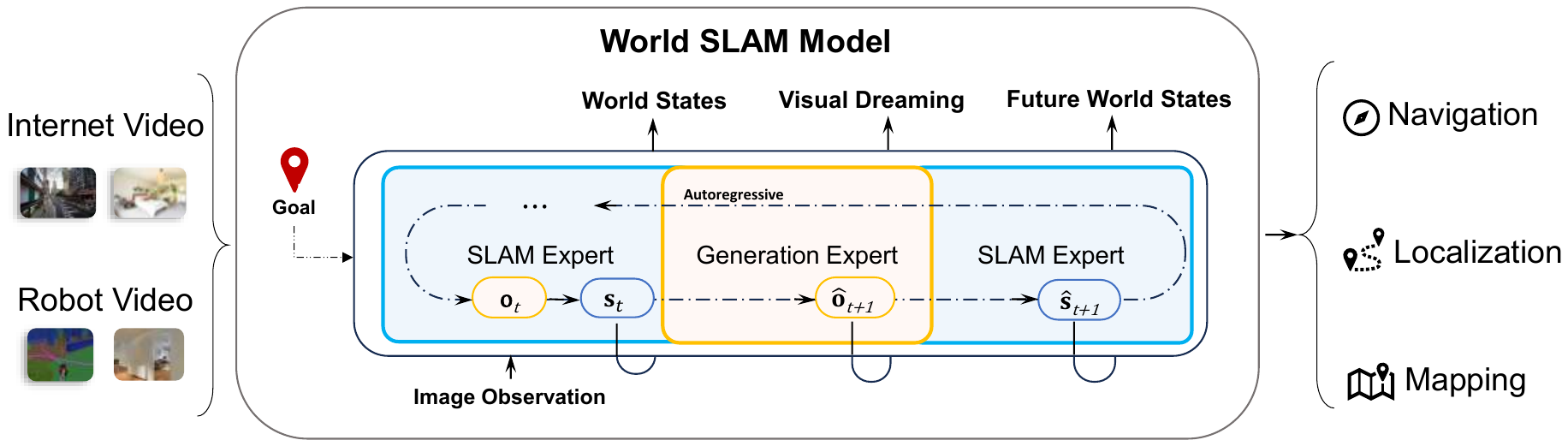}
    \caption{
    \textbf{Overview of the World SLAM Model.} WSM integrates a SLAM expert and a generation expert into a unified model that autoregressively estimates world states (i.e., depth and camera pose), dreams future observations, and infers future world states, supporting navigation, localization, and mapping. 
    }
    \label{fig:1}
\end{figure}

Our contributions are summarized as follows:
\begin{itemize}
    \item We revisit the role of SLAM in embodied systems and argue that
  downstream tasks can benefit more from incorporating the \emph{SLAM mechanism itself} into the model than from only consuming the outputs of an external SLAM module, such as poses, maps or tokens.

    \item We introduce \textbf{World SLAM Model (WSM)}, which incorporates
    incremental processing, persistent memory, spatial estimation, and
    closed-loop interaction into a unified navigation model.

    \item We evaluate WSM on both navigation and SLAM, demonstrating the
    potential of using SLAM as an intrinsic computational mechanism for
    downstream embodied tasks.
\end{itemize}
\section{Related Work}
\subsection{SLAM and Feed-Forward 3D Reconstruction}
Optimization-based SLAM systems~\citep{campos2021orb,teed2021droid} estimate camera poses and scene geometry from sequential observations, combining front--end tracking with loop closure and backend optimization to maintain global consistency and correct accumulated drift. Feed-forward reconstruction models instead regress geometry directly from images. VGGT~\citep{wang2025vggt}, D4RT~\citep{zhang2026d4rt} and $\pi^3$~\citep{wang2025pi} jointly predict camera parameters, depth maps, and point maps. StreamVGGT~\citep{zhuo2026streaming} adapts it to streaming inputs through causal attention with cached historical tokens. LingBot-Map~\citep{chen2026geometric} maintains a compact geometric context for long-sequence streaming reconstruction, and SLAM-Former~\citep{yuan2025slamformer} integrates incremental tracking, mapping, and global refinement into a single transformer. Across these formulations, however, geometric state estimation remains the primary objective, while navigation is left as a separate downstream task.

WSM takes a complementary direction: rather than treating SLAM solely as a geometric estimation objective, it incorporates the underlying SLAM mechanisms into a world-modeling framework for navigation.

\subsection{Visual Navigation}
Classical navigation systems follow a modular pipeline in which SLAM provides localization and mapping to a downstream planner and controller~\citep{marder2010office}. End-to-end methods~\citep{yang2023iplanner,cai2025navdp,roth2024viplanner} map observations directly to trajectories, reducing error propagation across module interfaces, but typically obtain the robot pose or goal position from external localization. Building on feed-forward 3D reconstruction models such as VGGT~\citep{wang2025vggt}, LoGoPlanner~\citep{peng2025logoplanner} fine-tunes a pretrained visual-geometry backbone with depth-derived scale priors for localization and scene reconstruction without relying on external odometry. However, these approaches still depend on high-fidelity depth sensing, a separately pretrained 3D reconstruction model, or complex, system-specific SLAM optimization modules. The mechanisms for geometric state estimation and drift correction are not embedded in the navigation model. 

In contrast, WSM brings the \emph{SLAM mechanism itself} into the navigation model: incremental processing with persistent memory maintains spatial context over time, while backend refinement mitigates accumulated drift.

\subsection{World Models}
World models capture how a world appears across views and evolves over time, allowing unobserved states to be generated, reconstructed, or simulated. Cosmos~\citep{agarwal2026cosmos} shows that world foundation models pretrained on large-scale video can be post-trained into customized world models for downstream physical AI applications, and Atlas~\citep{worldlabs2026atlas} indicates that grounding views at explicit camera poses allows a single model to support generation, reconstruction, and simulation. For downstream navigation and manipulation tasks, NWM~\citep{11093950} shows that an action-conditioned video model can serve as a navigation planner that incorporates new constraints at planning time, unlike supervised policies with fixed behavior. World action models such as LingBot-VA~\citep{li2026lingbotva} and Fast-WAM~\citep{yuan2026fastwam} jointly model future visual observations and actions for embodied control. Existing world models have explored predictive visual dynamics and explicit geometric representations, but they generally do not use a shared, persistent metric state to connect visual prediction with downstream tasks. 

WSM instead adopts the SLAM paradigm for world modeling, combining incremental state updates with backend refinement to maintain a spatially consistent world state over long-horizon interaction.

\section{World SLAM Model}
\subsection{Overview}
We formulate the navigation-oriented world SLAM model in terms of (1) observation-conditioned visual dynamics prediction, (2) observation-conditioned metric state estimation, and (3) dream-conditioned trajectory recovery. Observations are encoded by the Wan2.2 causal VAE~\citep{wan2025} with a temporal compression ratio of $M=4$, so each latent frame covers $M$ raw frames. At latent timestep $t$, the model receives RGB observations $\mathbf{o}_t\in\mathcal{O}$ and represents their metric state as $\mathbf{s}_t=(\mathbf{d}_t,\{\mathbf{g}_{t,m}\}_{m=1}^{M})$, where $\mathbf{d}_t\in\mathcal{D}$ denotes scene depth and $\mathbf{g}_{t,m}\in\mathcal{G}$ denotes the camera pose of the $m$-th raw frame. The model operates on chunks of $K$ timesteps, and we use $\mathbf{O}_n$ and $\mathbf{S}_n$ to denote the observations and metric states of the $n$-th chunk.

\begin{figure}[t] 
    \centering
    \includegraphics[width=\columnwidth]{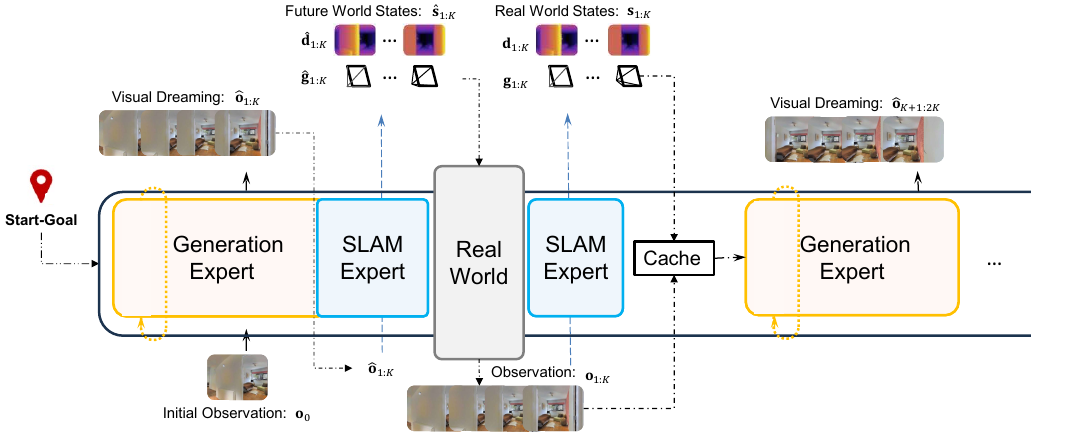} 
    \caption{\textbf{Framework overview.} Our model unifies visual dreaming and SLAM within an autoregressive \emph{Mixture-of-Transformers (MoT)} for start--goal navigation.
    Conditioned on the start--goal and the initial observation $\mathbf{o}_0$, the generation expert first \emph{dreams} future RGB observations $\hat{\mathbf{o}}_{1:K}$, from which the SLAM expert recovers future world states $\hat{\mathbf{s}}_{1:K}$ as an \emph{inverse dynamics model}.
    After execution, the SLAM expert estimates real world states $\mathbf{s}_{1:K}$ from the received observations $\mathbf{o}_{1:K}$, which are cached to condition the next dream $\hat{\mathbf{o}}_{K+1:2K}$.}
    \label{fig:architecture}
\end{figure}

\paragraph{Method overview.} Figure~\ref{fig:architecture} illustrates our framework.
Section~\ref{sec:method:video-slam} introduces our navigation formulation with the video model and the SLAM model. Section~\ref{sec:method:arch} presents our unified architecture and its autoregressive modeling.
Section~\ref{sec:method:train} describes pretraining and post-training, and Section~\ref{sec:method:infer} describes our inference pipeline.

\subsection{Video Model and SLAM Model for Navigation}
\label{sec:method:video-slam}
For clarity, we formulate each step at the granularity of one latent frame; in practice, the model processes a chunk of $K$ latent frames jointly, \emph{e.g.}, dreaming $\hat{\mathbf{o}}_{t+1:t+K}$ at once.

\paragraph{Dream for visual states.}
Given RGB observations $\mathbf{o}_{\leq t}$ and their estimated metric states $\mathbf{s}_{\leq t}$, the visual dynamics model $p_\theta$ dreams the future visual state:
\begin{equation}
\label{eq:dream}
    \hat{\mathbf{o}}_{t+1}
    \sim p_\theta\!\left(\cdot \mid \mathbf{o}_{\leq t},\mathbf{s}_{\leq t}\right),
\end{equation}
where the hat denotes dreamed quantities. The dreamed RGB sequence describes a visual transition but does not itself specify the corresponding metric motion. We recover that motion through the SLAM expert below.

\paragraph{SLAM for world states.}

The SLAM expert $f_\psi$ recovers camera motion and scene geometry from temporal visual observations. Applied to a dreamed visual transition, it infers the corresponding camera motion, which amounts to inverse dynamics. The same $f_\psi$ operates on observed and dreamed sequences:
\begin{equation}
\label{eq:past_idm}
    \mathbf{s}_{t+1}
    =f_\psi\!\left(\mathbf{o}_{t+1}\mid
    \mathbf{o}_{\leq t},\mathbf{s}_{\leq t}\right),
\end{equation}
\begin{equation}
\label{eq:future_idm}
    \hat{\mathbf{s}}_{t+1}
    =f_\psi\!\left(\hat{\mathbf{o}}_{t+1}\mid
    \mathbf{o}_{\leq t},\mathbf{s}_{\leq t}\right).
\end{equation}

To correct drift accumulated in incremental estimation, the SLAM expert runs a local backend:
\begin{equation}
\label{eq:local_backend}
    \mathbf{s}_{t-k+1:t}
    =f_\psi\!\left(\mathbf{o}_{t-k+1:t}\mid
    \mathbf{o}_{\leq t-k},\mathbf{s}_{\leq t-k}\right),
    \quad k\leq K,
\end{equation}
which jointly re-estimates the latest $k$ frames within a chunk of size $K$, and updates their states and KV cache.

\paragraph{Start-goal navigation.}
We additionally condition the dreaming process on the goal $\mathbf{p}_{\mathrm{goal}}\in\mathbb{R}^{3}$, a position in the first-camera coordinate frame without orientation:
\begin{equation}
\label{eq:goal_based_dream}
    \hat{\mathbf{o}}_{t+1}
    \sim p_\theta\!\left(\cdot\mid
    \mathbf{o}_{\leq t},\mathbf{s}_{\leq t},\mathbf{p}_{\mathrm{goal}}\right).
\end{equation}
The pose sequence recovered from this goal-conditioned dream by~\cref{eq:future_idm} serves as the metric plan.

\subsection{Unified Architecture and Autoregressive Modeling}
\label{sec:method:arch}

\paragraph{Architecture.}
We instantiate $p_\theta$ and $f_\psi$ within a single Mixture-of-Transformers (MoT)~\citep{liang2025mixtureoftransformers} with three experts: a generation expert for RGB tokens, and two experts for metric states, depth and pose, which together form the SLAM expert. Each expert keeps its own parameters, while all tokens interact through joint self-attention at every layer. RGB observations and depth are encoded by the Wan2.2 VAE and patchified into tokens, and pose tokens are decoded into camera poses by a pose head. Since depth and pose tokens are mutually visible and regressed jointly, we refer to the depth and pose experts together as the SLAM expert $f_\psi$ (Figure~\ref{fig:architecture}). The goal $\mathbf{p}_{\mathrm{goal}}$ is embedded as a goal token that conditions only the generation expert via cross-attention.

\begin{figure}[t]
    \centering
    \includegraphics[width=\linewidth]{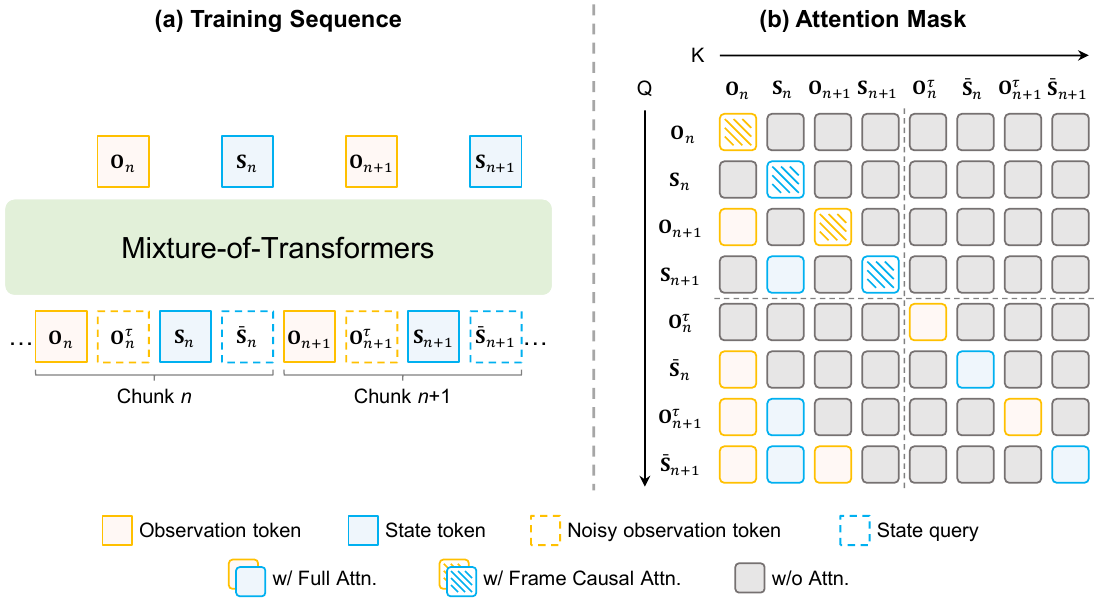}
    \caption{\textbf{Training sequence and attention mask.}
    (a) Each chunk $n$ consists of clean tokens $\mathbf{O}_n, \mathbf{S}_n$ and prediction tokens $\mathbf{O}^{\tau}_n, \bar{\mathbf{S}}_n$, and all chunks of an episode are processed by the Mixture-of-Transformers in a single forward pass.
    (b) Attention mask. Temporal order is preserved across chunks, and observation and state tokens in the history context are invisible to each other. No prediction token can attend to its own target.}
    \label{fig:attn_mask}
\end{figure}

\paragraph{Training sequence.}
Following the teacher-forcing scheme of LingBot-VA~\citep{li2026lingbotva}, we organize the input tokens of each chunk $n$ into two groups (Figure~\ref{fig:attn_mask}a): clean tokens $\mathbf{O}_n$ and $\mathbf{S}_n$, which serve as context; and prediction tokens, which comprise noisy RGB tokens $\mathbf{O}^{\tau}_n$ for the generation expert and fixed state queries $\bar{\mathbf{S}}_n$ with zero depth and identity pose for the SLAM expert. All chunks of an episode are thus supervised in parallel within a single forward pass.

\paragraph{Attention mask.}
The mask (Figure~\ref{fig:attn_mask}b) preserves temporal order across chunks while allowing joint prediction within a chunk. Clean tokens attend frame-causally to clean tokens of the same type (RGB or state), which keeps the RGB and state histories decoupled. In chunk $n$, noisy RGB tokens attend to the clean tokens of all earlier chunks and bidirectionally to one another; the past metric states help the generation expert imagine futures that are geometrically consistent with the observed scene. State queries share this visibility and additionally attend to the clean RGB tokens $\mathbf{O}_n$, since the metric states of a chunk are estimated from its own frames. Noisy RGB tokens and state queries do not attend to each other, which prevents information leakage.

\subsection{Training}
\label{sec:method:train}

\paragraph{Pretraining: joint teacher-forcing training.}
We train RGB generation with flow matching~\citep{lipman2022flow} and supervise depth and camera poses by regression. Given Gaussian noise $\boldsymbol{\epsilon}\sim\mathcal{N}(\mathbf{0},\mathbf{I})$ and a flow time $\tau\in[0,1]$, we construct $\mathbf{O}^{\tau}_n=(1-\tau)\mathbf{O}_n+\tau\boldsymbol{\epsilon}$ and train the model to predict the corresponding velocity:
\begin{equation}
    \mathcal{L}_{\mathrm{FM}}
    =\mathbb{E}_{\mathbf{O}_n,\boldsymbol{\epsilon},\tau}\!\left[
      \left\|
      v_\theta(\mathbf{O}^{\tau}_n,\tau\mid\mathbf{c}_n)
      -(\boldsymbol{\epsilon}-\mathbf{O}_n)
      \right\|_2^2
    \right],
    \label{eq:flow_matching}
\end{equation}
where $\mathbf{c}_n$ contains the goal token and the tokens visible to $\mathbf{O}^{\tau}_n$ under the attention mask.  The SLAM expert instead regresses the clean depth latents and metric camera poses of chunk $n$ directly from the state queries. All poses are expressed relative to the first camera. The overall objective is
\begin{equation}
    \mathcal{L}_{\mathrm{pre}}
    =\mathcal{L}_{\mathrm{FM}}
    +\lambda_d\mathcal{L}_{d}
    +\lambda_p\left(\mathcal{L}_{\mathrm{abs}}+\mathcal{L}_{\mathrm{rel}}\right),
    \label{eq:stage1_loss}
\end{equation}
where $\mathcal{L}_{d}$ is the mean squared error on depth latents, $\mathcal{L}_{\mathrm{abs}}$ and $\mathcal{L}_{\mathrm{rel}}$ are smooth-$L_1$ losses on absolute and pairwise relative poses, and $\lambda_d$ and $\lambda_p$ balance the terms. Details are provided in~\cref{app:pretrain_loss}.

\paragraph{Post-training: safety-prioritized reinforcement learning.}
Imitation pretraining suffers from covariate shift at deployment, and stochastic sampling can yield futures that collide with obstacles. Since the SLAM expert turns each generated video into a metric trajectory, generated futures can be scored against the occupancy maps and expert trajectories of the dataset, without a simulator. We therefore post-train the generation expert with DiffusionNFT~\citep{zheng2026diffusionnft}: for each history rolled out by the model itself and goal, a group of $G$ RGB rollouts is sampled, and the SLAM expert recovers their trajectories $\boldsymbol{\xi}_{1:G}$. We adapt the reward of CompassNav~\citep{li2026compassnav} with a safety gate, so that safety takes strict priority:
\begin{equation}
    r_i=\frac{s_i\exp(-C_i/T)}{\sum_{j=1}^{G}s_j\exp(-C_j/T)},
    \label{eq:reward}
\end{equation}
where $r_i$ is the reward of the $i$-th rollout, $s_i\in\{0,1\}$ indicates whether $\boldsymbol{\xi}_i$ is collision-free and $T$ is a temperature. The cost $C_i$ is the mean squared error between the per-step geodesic progress toward the goal along $\boldsymbol{\xi}_i$ and along the expert trajectory. Measuring progress rather than spatial deviation rewards safe routes that approach the goal at the expert's pace, even when they depart from the expert path. Only the self-attention LoRA~\citep{hu2021lora} adapters of the generation expert are updated; the SLAM expert stays frozen so the reward cannot be increased by updating the trajectory estimator itself. Details are provided in the Appendix~\ref{app:nft}.

\subsection{Autoregressive Closed-Loop Inference}
\label{sec:method:infer}
The model replans as new observations arrive after execution. At each planning step, it generates a goal-conditioned RGB sequence, from which the SLAM expert directly regresses future depth and poses in a single forward pass per chunk, without iterative denoising. A low-level controller tracks a short prefix of the pose trajectory, after which the newly observed frames are appended to the history. The local backend then re-estimates the depth and poses of all observed frames in the current chunk and replaces their previous estimates, while their RGB tokens remain unchanged in the KV cache. Earlier estimates in a chunk are thus refined as more of its frames arrive, and remain fixed once the chunk is complete. Generated frames are discarded after each planning step, so the persistent history contains only observed frames and their estimated depth and poses. The next prediction uses this updated history and reuses the KV cache across planning steps.

\section{Experimental Evaluation}
\label{sec:experiments}

We evaluate WSM on closed-loop start--goal navigation (Section~\ref{sec:navigation_results}) and on SLAM, measuring camera pose estimation and dense reconstruction on held-out trajectories (Section~\ref{sec:slam_results}). Section~\ref{sec:ablation} ablates two components: the SLAM expert, by comparing WSM with IDM (w/o SLAM), which decodes camera motion from dreamed frames without estimating depth or localizing observed frames, and the local backend. Appendix~\ref{app:stage_ablation} compares the model before and after post-training.

\subsection{Experimental Setup}
\label{sec:exp_setup}

\paragraph{Training data and details.}
We train on InternVLN--N1~\citep{internvla-n1} datasets and reserve 5\% of the trajectories as the SLAM test split. These test trajectories are excluded from both pre-training and post-training. All experts are initialized from Wan2.2-5B~\citep{wan2025}. The generation expert directly inherits the pretrained DiT weights with a hidden width of 3072, while the depth and pose experts are initialized by linearly interpolating the pretrained weights to hidden widths of 1536 and 768, respectively LingBot-VA. All training is conducted on 48 A100 GPUs. Teacher-forcing pretraining runs for 10 epochs with a global batch size of 48, and the chunk size is randomly sampled from 1 to 4 latent frames. We use AdamW with a weight decay of 0.01, linearly warming up the learning rate to $3\times10^{-5}$ over the first 5\% of steps and then decaying it to $3\times10^{-6}$ with a cosine schedule. Post-training runs for one epoch with the same learning rates, and a weight decay of $10^{-4}$. More details are provided in Appendix~\ref{app:exp}.

\paragraph{Inference details.}
We generate $K=4$ future RGB latents at $320\times192$ resolution using 8 sampling steps. Each latent frame yields $M=4$ poses, giving 16 predicted poses per planning step. The controller executes to the fourth predicted pose before replanning from four new RGB observations, and the KV cache retains up to 128 observed latent frames. All inference is performed on a single NVIDIA RTX 4090 GPU.

\subsection{Navigation Evaluation}
\label{sec:navigation_results}

Following LoGoPlanner~\citep{peng2025logoplanner}, we evaluate WSM on the InternVLA--N1~\citep{internvla-n1} start--goal benchmark in a closed-loop setting. The benchmark comprises 40 InternScenes environments (20 Home and 20 Commercial) with 100 start--goal episodes each, all disjoint from training datasets.

The goal of each episode is specified as a position in the robot's initial frame and remains fixed as the robot moves. The model therefore has to track its own pose relative to the initial frame and decide when to stop. WSM takes only RGB observations as input and estimates camera poses and depth with its SLAM expert, without external odometry or post-processing. Following the benchmark protocol, an episode is successful if the robot decelerates or stops within $1$~m of the goal. We report Success Rate (SR) and Success weighted by Path Length (SPL).

Table~\ref{tab:sim_results} compares WSM with DD-PPO~\citep{wijmans2019dd}, iPlanner~\citep{yang2023iplanner}, ViPlanner~\citep{roth2024viplanner}, and LoGoPlanner. For LoGoPlanner, we report both its published results and our re-evaluation with its latest official code.

\begin{table}[t]
\caption{\textbf{Navigation results on the InternVLA-N1 start--goal navigation benchmark.}
SR and SPL are reported in percent. Checkmarks indicate the input modalities. Baseline results are taken from LoGoPlanner. $^\dagger$ means our re-evaluation with its latest official code.}
\label{tab:sim_results}
\centering
\definecolor{rowcolor}{rgb}{0.898,0.949,0.969}
\begin{tabular}{lcccccc}
\toprule
\multirow{2}{*}{Method}
& \multicolumn{2}{c}{Input}
& \multicolumn{2}{c}{Home}
& \multicolumn{2}{c}{Commercial} \\
\cmidrule(lr){2-3} \cmidrule(lr){4-5} \cmidrule(lr){6-7}
& RGB & Depth
& SR$\uparrow$ & SPL$\uparrow$
& SR$\uparrow$ & SPL$\uparrow$ \\
\midrule
DD-PPO~\citep{wijmans2019dd}
& $\checkmark$ & $\checkmark$
& 0.4 & 0.4 & 5.3 & 5.2 \\
iPlanner~\citep{yang2023iplanner}
& -- & $\checkmark$
& 43.0 & 40.6 & 54.6 & 52.8 \\
ViPlanner~\citep{roth2024viplanner}
& $\checkmark$ & $\checkmark$
& 45.0 & 43.2 & 63.7 & 61.9 \\
LoGoPlanner~\citep{peng2025logoplanner}
& $\checkmark$ & $\checkmark$
& 57.3 & 52.4 & 67.1 & 63.9 \\
\midrule
LoGoPlanner$^\dagger$~\citep{peng2025logoplanner}
& $\checkmark$ & $\checkmark$
& 59.6 & 53.6 & 65.8 & 62.0 \\
\rowcolor{rowcolor}
\textbf{World SLAM Model}
& $\checkmark$ & --
& \textbf{70.3} & \textbf{66.3}
& \textbf{73.9} & \textbf{71.8} \\
\bottomrule
\end{tabular}
\end{table}

\begin{figure}[t]
    \centering
    \includegraphics[width=\linewidth]{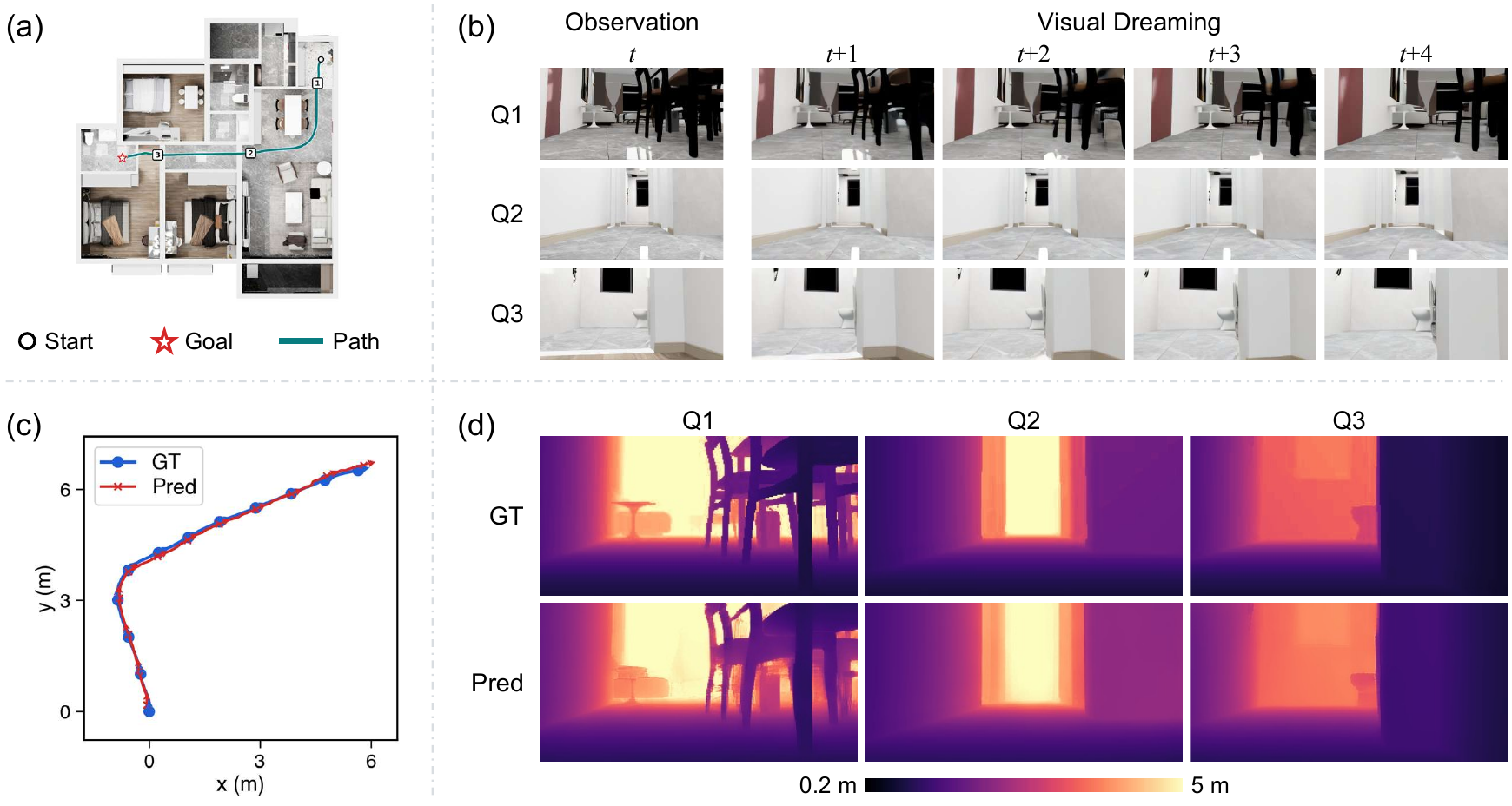}
    \caption{\textbf{Visualization of navigation results in simulation.}
    (a) Navigation result; Q1--Q3 mark the queries shown in (b) and (d).
    (b) RGB observation at time $t$ and the frames dreamed by WSM for $t{+}1$ to $t{+}4$ at each query.
    (c) Camera trajectory estimated by the SLAM expert vs.\ ground truth.
    (d) Ground-truth and predicted depth at time $t$ for each query, colored from 0.2 to 5\,m.}
    \label{fig:navigation_visualization}
\end{figure}

WSM achieves the highest SR in both Home and Commercial environments, improving over LoGoPlanner by 10.7 and 8.1 percentage points, respectively. As shown in Figure ~\ref{fig:navigation_visualization}, WSM reaches the goal in an unseen Home scene, and its estimated camera trajectory and predicted depth closely match the ground truth. We attribute this to jointly modeling visual dreaming and SLAM in a single model, where localization and mapping from RGB observations provide the pose and geometry required for goal reaching without external odometry.

\subsection{SLAM Evaluation}
\label{sec:slam_results}

We evaluate camera pose estimation and dense 3D reconstruction on all RGB trajectories in the 5\% InternVLN-N1 test split and compare WSM with SLAM-Former~\citep{yuan2025slamformer}. For the comparison reported here, WSM is trained on the InternVLN-N1 navigation dataset alone, whereas SLAM-Former starts from a model pretrained on large-scale 3D datasets and is then fine-tuned on the same InternVLN-N1 training split. Both the released SLAM-Former checkpoint and its fine-tuned version are evaluated on the 5\% test split.

The input resolutions are $320\times192$ for WSM and $518\times518$ for SLAM-Former. Camera pose estimation is measured by camera-center ATE RMSE after per-sequence $\mathrm{Sim}(3)$ alignment, while dense reconstruction is evaluated by accuracy (Acc.), the RMSE of the distances from predicted points to their nearest ground-truth points, after alignment and rigid ICP refinement.
\begin{table}[t]
\caption{\textbf{SLAM evaluation on the 5\% InternVLN-N1 test split.}
$^\dagger$ denotes the released checkpoint; $^\ddagger$ denotes the model fine-tuned on the 95\% InternVLN-N1 training split.}
\label{tab:slam_results}
\centering
\definecolor{rowcolor}{rgb}{0.898,0.949,0.969}
\begin{tabular}{lcc}
\toprule
Method & ATE RMSE (m)$\downarrow$ & Acc. RMSE (m)$\downarrow$ \\
\midrule
SLAM-Former$^\dagger$~\citep{yuan2025slamformer}
& 0.051 & 0.057 \\
SLAM-Former$^\ddagger$~\citep{yuan2025slamformer}
& 0.041 & \textbf{0.041} \\
\rowcolor{rowcolor}
\textbf{World SLAM Model}
& \textbf{0.021} & 0.062 \\
\bottomrule
\end{tabular}
\end{table}

As shown in Table~\ref{tab:slam_results}, WSM achieves an ATE RMSE of $0.021$~m, compared with $0.041$~m for the fine-tuned SLAM-Former. For dense reconstruction, WSM obtains an Acc. RMSE of $0.062$~m, which is worse than the $0.041$~m of the fine-tuned SLAM-Former. This gap may result from encoding depth in the appearance-oriented VAE latent space.

\subsection{Ablation Studies}
\label{sec:ablation}

\paragraph{SLAM IDM.}
Existing world action models~\citep{li2026lingbotva} dream future frames and decode actions from the dreamed transitions through inverse dynamics. IDM (w/o SLAM) follows this design: it removes the depth expert and all depth and pose estimation for observed frames, and its pose expert only performs inverse dynamics, generating future camera poses from the observed and dreamed RGB, conditioned on the goal and on previously issued poses. Both models share the same initialization and training data and are evaluated after pretraining. 

As shown in Table~\ref{tab:ablation_stage1}, IDM (w/o SLAM) trails WSM without post-training by about 36 SR points in both Home and Commercial. This gap likely reflects IDM’s less reliable localization and scene-geometry estimation, which hinder planning toward a goal fixed in the first-camera frame. WSM uses a shared SLAM expert to localize and map observed frames and perform inverse dynamics on dreamed frames in the same metric coordinates, keeping the plan anchored to the goal.

\paragraph{Local backend.}
The local backend in \cref{eq:local_backend} refines previous states. We compare it with incremental estimation alone, where each state is estimated once by \cref{eq:past_idm} and then kept fixed.
Both variants are post-trained from the same pretrained model, each using its own state update in both post-training and evaluation.
As shown in Table~\ref{tab:ablation_local_backend}, the local backend improves SR/SPL by 2.0/1.1 on Home and by 1.3/1.3 on Commercial, indicating that refining recent states reduces error accumulation in the history and leads to more reliable closed-loop navigation.

\begin{table}[t]
\caption{\textbf{SLAM IDM ablation.}
Both models are evaluated after pretraining, without post-training.
SR and SPL are reported in percent.}
\label{tab:ablation_stage1}
\centering
\definecolor{rowcolor}{rgb}{0.898,0.949,0.969}
\begin{tabular}{lcccc}
\toprule
\multirow{2}{*}{Method}
& \multicolumn{2}{c}{Home}
& \multicolumn{2}{c}{Commercial} \\
\cmidrule(lr){2-3} \cmidrule(lr){4-5}
& SR$\uparrow$ & SPL$\uparrow$
& SR$\uparrow$ & SPL$\uparrow$ \\
\midrule
IDM (w/o SLAM)
& 28.1 & 27.1 & 32.1 & 31.2 \\
\rowcolor{rowcolor}
\textbf{World SLAM Model (w/o post-training)}
& \textbf{64.6} & \textbf{61.5} & \textbf{68.5} & \textbf{66.9} \\
\bottomrule
\end{tabular}
\end{table}

\begin{table}[t]
\caption{\textbf{Local backend ablation.}
Both variants are post-trained from the same pretrained model and differ only in whether the local backend is used in history rollout, during both post-training and closed-loop evaluation.
SR and SPL are reported in percent.}
\label{tab:ablation_local_backend}
\centering
\definecolor{rowcolor}{rgb}{0.898,0.949,0.969}
\begin{tabular}{lcccc}
\toprule
\multirow{2}{*}{Method}
& \multicolumn{2}{c}{Home}
& \multicolumn{2}{c}{Commercial} \\
\cmidrule(lr){2-3} \cmidrule(lr){4-5}
& SR$\uparrow$ & SPL$\uparrow$
& SR$\uparrow$ & SPL$\uparrow$ \\
\midrule
w/o local backend
& 68.3 & 65.2 & 72.6 & 70.5 \\
\rowcolor{rowcolor}
\textbf{w/ local backend}
& \textbf{70.3} & \textbf{66.3} & \textbf{73.9} & \textbf{71.8} \\
\bottomrule
\end{tabular}
\end{table}

\section{Conclusion}
\label{sec:conclusion}
We presented the World SLAM Model, in which the generation expert dreams goal-conditioned futures and the SLAM expert estimates world states from both observed and dreamed frames, maintaining one persistent world state in a closed loop. Using RGB input only, WSM achieves the highest SR and SPL on start-goal navigation among the evaluated methods while retaining accurate spatial estimation. These results indicate that navigation benefits not only from what SLAM estimates but also from how it maintains and refines a world state, highlighting SLAM as an intrinsic mechanism for long-horizon world modeling and embodied interaction.

WSM encodes depth in the Wan2.2 VAE latent space, which is designed for appearance rather than geometry, trading geometric fidelity for compatibility with the pretrained video generator. Its history KV cache is also kept without compression, so memory and computation grow with the number of observed frames, limiting long-horizon deployment. Future work includes geometry-native latent representations for depth and efficient history compression.

\newpage
\subsection*{AI use statement}

In this work, we used generative AI tools for {none of the required disclosure}.
We have not used generative AI tools for {all of the required disclosure}, 
and {all of the required disclosure} are not applicable to this work.
Additionally, we used generative AI tools for {polishing paper writing}. We have reviewed all AI-assisted work. 
We use AI tools to correct the input sentence grammar and refine the writing to be formal and readable.
We take responsibility for the final content of this work,
including text, claims or artifacts produced with the aid of generative AI.

\bibliography{iclr2027_conference}
\bibliographystyle{iclr2027_conference}

\clearpage
\appendix
\appendix
\section{Method Details}
\label{app:method}

\subsection{Pretraining Objective}
\label{app:pretrain_loss}

\paragraph{RGB generation.}
The RGB loss follows eq.~(\ref{eq:flow_matching}). The flow time is sampled per chunk and
shifted as $\tau=\gamma\sigma/(1+(\gamma-1)\sigma)$ with $\sigma\sim\mathcal{U}(0,1)$,
and the loss is weighted by a bell-shaped function of $\tau$. The first chunk of
each training sequence serves as a clean visual anchor and is excluded from the loss.

\paragraph{Depth.}
Metric depth $d$ is clipped to $[d_{\min},d_{\max}]=[0.2,5]$\,m and normalized in
log space,
\begin{equation}
u=\frac{\log(d/d_{\min})}{\log(d_{\max}/d_{\min})}\in[0,1].
\end{equation}
The map $2u-1$ is replicated to three channels and encoded by the Wan2.2 VAE, and
$\mathcal{L}_d$ is the mean squared error on the resulting depth latents after
per-channel standardization. At inference, metric depth is recovered by decoding
the predicted latents and inverting the log mapping.

\paragraph{Camera pose.}
Following SLAM-Former~\citep{yuan2025slamformer}, we supervise camera poses with
the smooth-$L_1$ (Huber) norm $\|\cdot\|_\delta$, applied to the translation and
the unit-quaternion rotation of a pose difference. Since WSM predicts metric poses
in the first-camera frame, no scale alignment is applied, and both absolute and
pairwise relative poses are supervised:
\begin{equation}
\mathcal{L}_{\mathrm{abs}}=\sum_{j}\big\|\hat{\mathbf{g}}_j-\mathbf{g}_j\big\|_\delta,
\qquad
\mathcal{L}_{\mathrm{rel}}=\sum_{i,j}\big\|\hat{\mathbf{g}}_i^{-1}\hat{\mathbf{g}}_j-\mathbf{g}_i^{-1}\mathbf{g}_j\big\|_\delta,
\end{equation}
where the sums run over all raw frames of a training sequence and are normalized
by the number of terms.

\subsection{Post-Training}
\label{app:nft}

\paragraph{Reward.}
For each dream $\hat{\mathbf{O}}^{(i)}$, the SLAM expert recovers $L=KM$ camera
positions $\mathbf{x}_{i,1:L}$, which are placed in the occupancy map through the
dataset pose $\mathbf{x}_0$ of the query frame. Let $D(\mathbf{x})$ be the geodesic
distance from $\mathbf{x}$ to the goal region of radius $r_g$, and let
$\Delta_{i,l}=D(\mathbf{x}_{i,l-1})-D(\mathbf{x}_{i,l})$ be the progress of step $l$,
with $\mathbf{x}_{i,0}=\mathbf{x}_0$. The cost in eq.~(\ref{eq:reward}) is
\begin{equation}
C_i=\frac{1}{L}\sum_{l=1}^{L}\big(\Delta_{i,l}-\Delta^{\star}_{l}\big)^2,
\end{equation}
where $\Delta^{\star}_{l}$ is the progress of the expert trajectory at the same
time step. The safety indicator $s_i=1$ if the robot footprint along the
trajectory stays in free space. In addition, the safe dream with the lowest cost
receives a bonus that grows with its cost margin over the second best.

\paragraph{Objective.}
Within each group, rewards are standardized into advantages $A_i$ and mapped to
optimality probabilities
\begin{equation}
\rho_i=\frac{1}{2}+\frac{1}{2}\,\mathrm{clip}\big(A_i/A_{\max},-1,1\big).
\end{equation}
Following DiffusionNFT~\citep{zheng2026diffusionnft}, let $v^{\mathrm{old}}$ be the
old policy that samples the dreams, an exponential moving average of the trained
policy $v_\theta$, and define the implicit positive and negative policies
$v^{+}=(1-\beta)v^{\mathrm{old}}+\beta v_\theta$ and
$v^{-}=(1+\beta)v^{\mathrm{old}}-\beta v_\theta$. For a dream noised as
$(1-\tau)\hat{\mathbf{O}}^{(i)}+\tau\epsilon$, we minimize
\begin{equation}
\mathcal{L}_{\mathrm{post}}=\mathbb{E}\Big[\rho_i\big\|v^{+}-(\epsilon-\hat{\mathbf{O}}^{(i)})\big\|_2^2
+(1-\rho_i)\big\|v^{-}-(\epsilon-\hat{\mathbf{O}}^{(i)})\big\|_2^2\Big]
+\lambda_{\mathrm{ref}}\big\|v_\theta-v^{\mathrm{ref}}\big\|_2^2,
\label{eq:app_post}
\end{equation}
where $v^{\mathrm{ref}}$ is the pretrained model, and the two error terms use the
adaptive normalization of DiffusionNFT. High-reward dreams pull $v_\theta$ toward
their flow-matching targets, and low-reward dreams push it away. The loss is
defined on the forward process and uses only the clean dreams, so it requires
neither likelihoods nor the sampling trajectory. Algorithm~\ref{alg:post}
summarizes the procedure.

\begin{algorithm}[t]
\caption{Post-training of WSM}
\label{alg:post}
\begin{algorithmic}[1]
\Require pretrained generation expert $v_\theta$ and SLAM expert $f_\psi$, queries $\mathcal{Q}$, group size $G$
\State Add LoRA adapters to the self-attention layers of $v_\theta$ and freeze all other parameters
\State $v^{\mathrm{old}}\gets v_\theta$
\For{each query in $\mathcal{Q}$}
  \State Sample $G$ dreams $\hat{\mathbf{O}}^{(1:G)}$ from $v^{\mathrm{old}}$ given the history and the goal
  \State Recover their trajectories with $f_\psi$
  \State Compute $s_i$, $C_i$, $r_i$, and $\rho_i$ for each dream
  \State Update the LoRA adapters by minimizing $\mathcal{L}_{\mathrm{post}}$ in eq.~(\ref{eq:app_post})
  \State Update $v^{\mathrm{old}}$ as a moving average of $v_\theta$
\EndFor
\end{algorithmic}
\end{algorithm}

\section{Experimental Details}
\label{app:exp}

\subsection{Implementation Details}
\label{app:impl}

\paragraph{Architecture.}
The three experts share 30 transformer layers with joint self-attention. The
generation expert keeps the width of Wan2.2-5B, with hidden size 3072, FFN size
14336, and 24 attention heads. The depth and pose experts have hidden sizes 1536
and 768 and FFN sizes 7168 and 3584, and are initialized by linearly interpolating
the pretrained weights to these widths. RGB and depth latents are patchified with patch size $(1,2,2)$.

\paragraph{Pretraining.}
Table~\ref{tab:app_hparams_pre} lists the pretraining hyperparameters. We set
$\lambda_d=1$, $\lambda_p=0.5$ in eq.~(\ref{eq:stage1_loss}). The
generation expert and the goal embedding use 0.3 times the base learning rate.

\paragraph{Post-training.}
Table~\ref{tab:app_hparams_post} lists the post-training hyperparameters. We post-train on about 10\% of the training trajectories, selected as hard examples. Starting
from the pretrained model, we add LoRA adapters with rank 32 and $\alpha=64$ to the
query, key, value, and output projections of the self-attention layers in the
generation expert. For each query, we sample $G=16$ dreams of $K=4$ latent frames
with 8 sampling steps and draw 4 flow times per dream. The reward uses
$T=0.01$\,m$^2$ and $r_g=0.2$\,m, and the objective uses $\beta=0.1$,
$A_{\max}=5$, and $\lambda_{\mathrm{ref}}=10^{-4}$.

\begin{table}[h]
\centering
\small
\caption{\textbf{Pretraining hyperparameters.}}
\label{tab:app_hparams_pre}
\begin{tabular}{ll}
\toprule
Hyperparameter & Value \\
\midrule
Initialization & Wan2.2-5B \\
Input resolution & $320\times192$ \\
Chunk size (latent frames) & uniform in $\{1,\dots,4\}$ \\
Timestep shift $\gamma$ & 5 \\
$\lambda_d$ / $\lambda_p$ / $\delta$ & 1 / 0.5 / 0.1 \\
Optimizer & AdamW ($\beta_1=0.9$, $\beta_2=0.95$, $\epsilon=10^{-8}$) \\
Weight decay & 0.01 \\
Peak / final learning rate & $3\times10^{-5}$ / $3\times10^{-6}$ \\
LR multiplier (generation expert, goal embedding) & 0.3 \\
Schedule & 5\% linear warmup, cosine decay \\
Gradient clipping & 2.0 \\
Epochs & 10 \\
Global batch size & 48 \\
Precision & bfloat16 \\
Hardware & 48 A100 GPUs \\
\bottomrule
\end{tabular}
\end{table}

\begin{table}[h]
\centering
\small
\caption{\textbf{Post-training hyperparameters.}}
\label{tab:app_hparams_post}
\begin{tabular}{ll}
\toprule
Hyperparameter & Value \\
\midrule
Initialization & pretrained model \\
Trainable parameters & self-attention LoRA of the generation expert \\
LoRA rank / $\alpha$ & 32 / 64 \\
Group size $G$ & 16 \\
Dreamed latent frames $K$ & 4 \\
Sampling steps & 8 \\
Flow times per dream & 4 \\
Timestep shift $\gamma$ & 5 \\
Temperature $T$ / goal radius $r_g$ & 0.01\,m$^2$ / 0.2\,m \\
$\beta$ / $A_{\max}$ / $\lambda_{\mathrm{ref}}$ & 0.1 / 5 / $10^{-4}$ \\
Optimizer & AdamW ($\beta_1=0.9$, $\beta_2=0.999$, $\epsilon=10^{-8}$) \\
Weight decay & $10^{-4}$ \\
Peak / final learning rate & $3\times10^{-5}$ / $3\times10^{-6}$ \\
Schedule & 3\% linear warmup, cosine decay \\
Gradient clipping & 1.0 \\
Epochs & 1 \\
Precision & bfloat16 \\
Hardware & 48 A100 GPUs \\
\bottomrule
\end{tabular}
\end{table}

\subsection{Two-Stage Ablation}
\label{app:stage_ablation}

\begin{figure}[h]
\centering
\includegraphics[width=\linewidth]{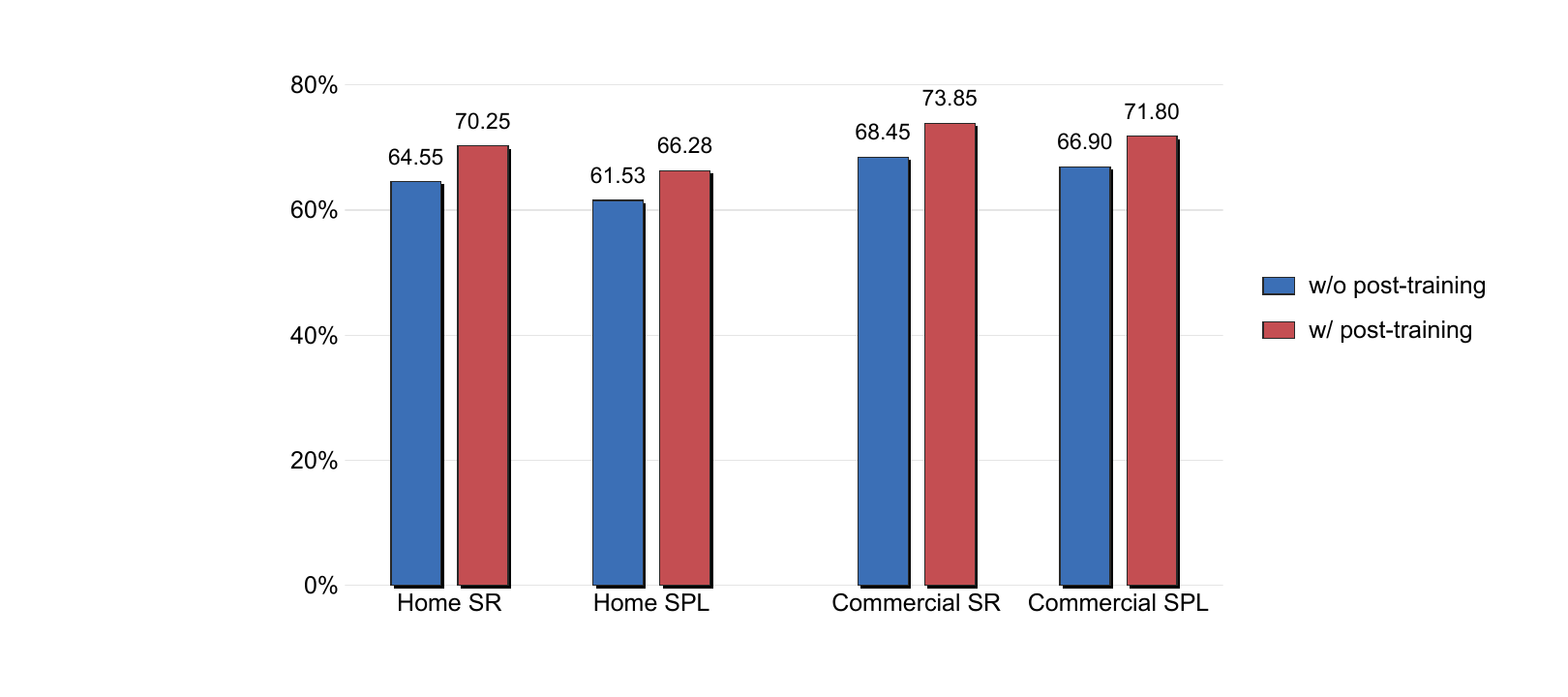}
\caption{\textbf{Two-stage ablation.} SR and SPL are reported in percent.}
\label{fig:app_stage}
\end{figure}

Figure~\ref{fig:app_stage} compares the pretrained model with the final model after post-training under the same inference settings. Post-training improves SR by 5.7 and 5.4 points and SPL by 4.8 and 4.9 points on Home and Commercial, respectively. Since only the generation expert is updated, the gains mainly reflect a change in which futures are dreamed: the reward favors dreams whose recovered trajectories are collision-free and approach the goal at the expert's pace, which teacher-forcing pretraining does not optimize explicitly.

\subsection{Additional Qualitative Results}
\label{app:qualitative}

\begin{figure}[p]
\centering
\includegraphics[width=\linewidth]{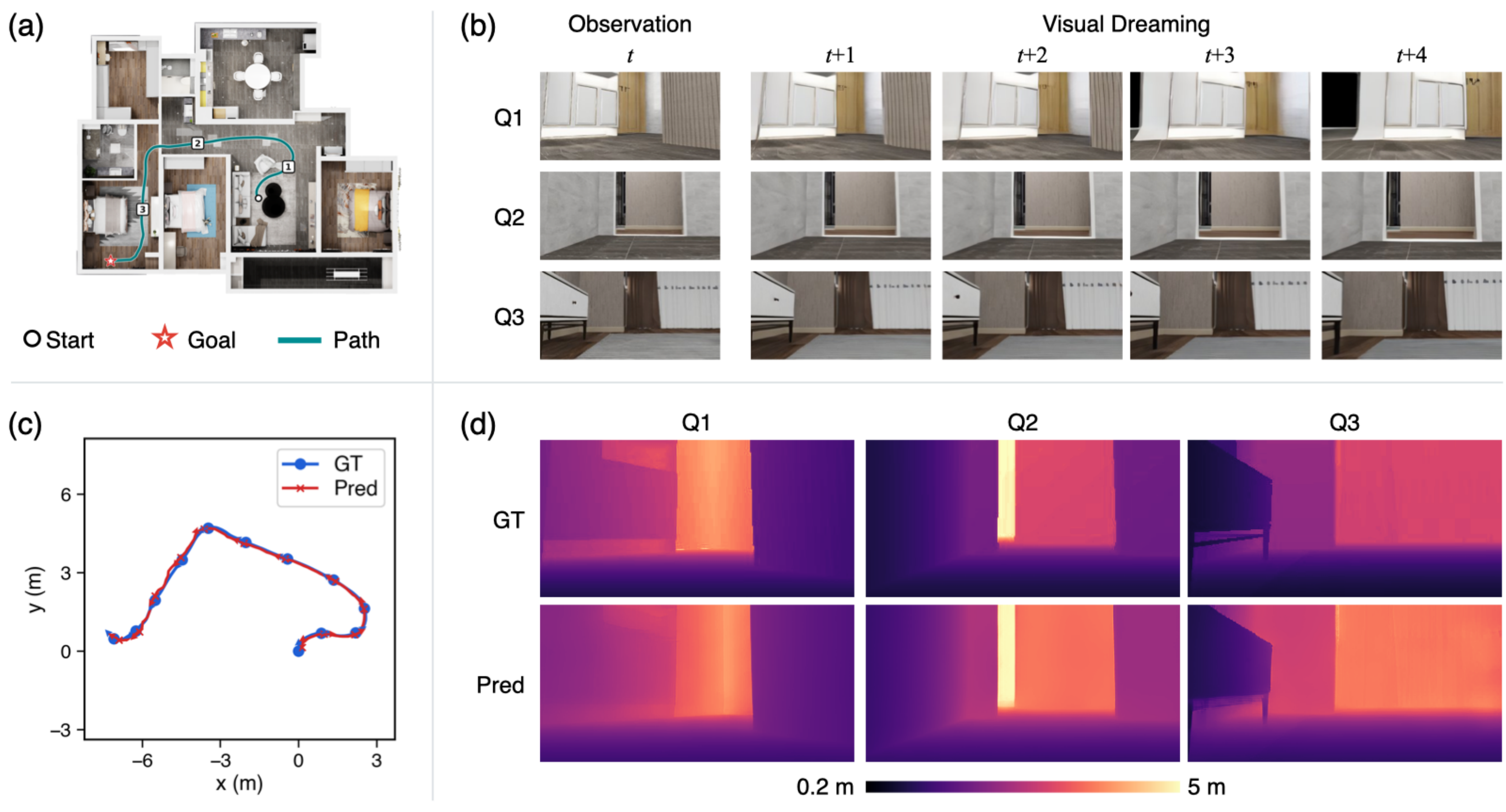}
\caption{\textbf{Navigation result in Home scene 15, episode 27.}}
\label{fig:app_vis_1}
\end{figure}

\begin{figure}[p]
\centering
\includegraphics[width=\linewidth]{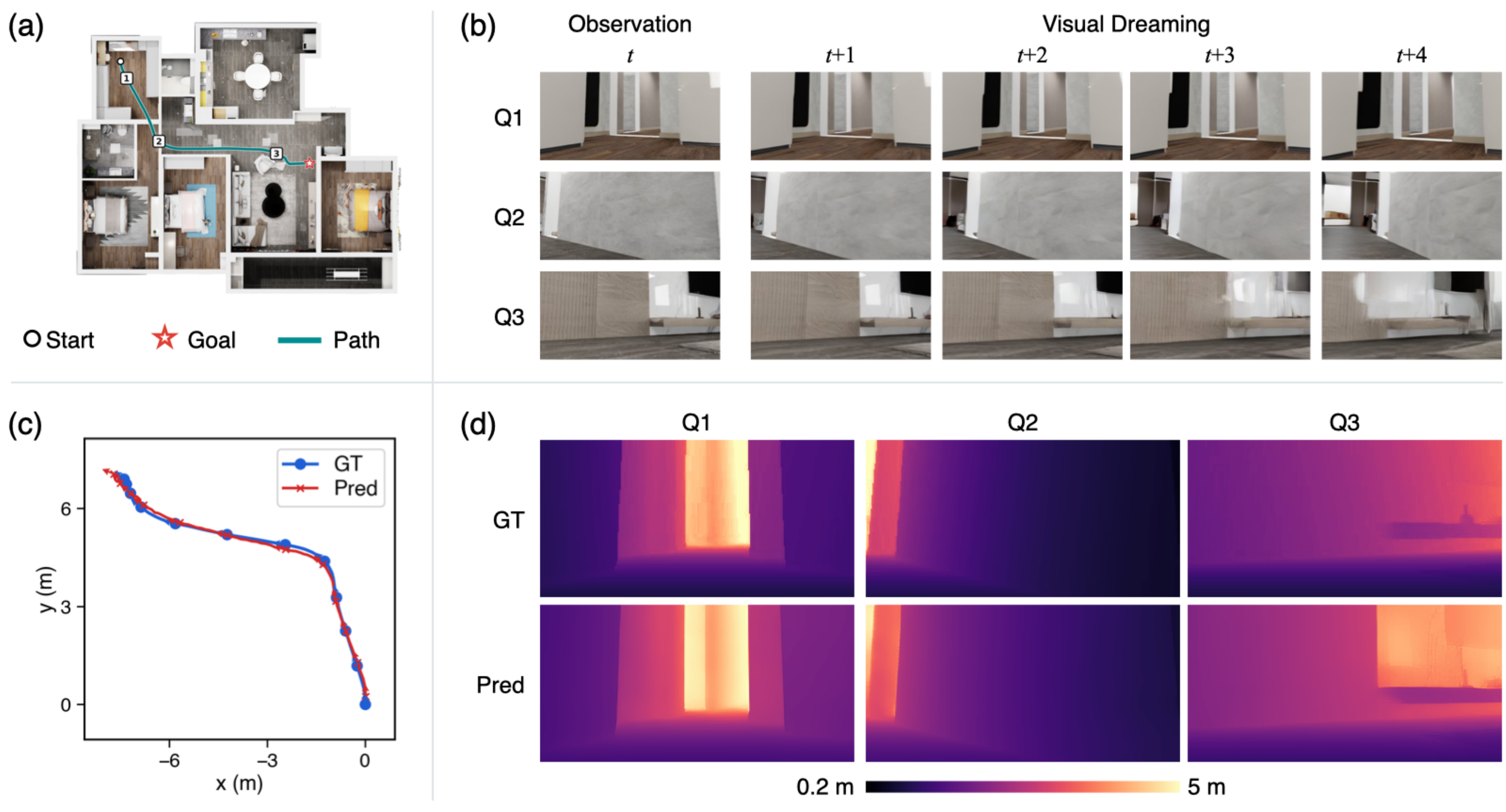}
\caption{\textbf{Navigation result in Home scene 15, episode 79.}}
\label{fig:app_vis_2}
\end{figure}

\begin{figure}[p]
\centering
\includegraphics[width=\linewidth]{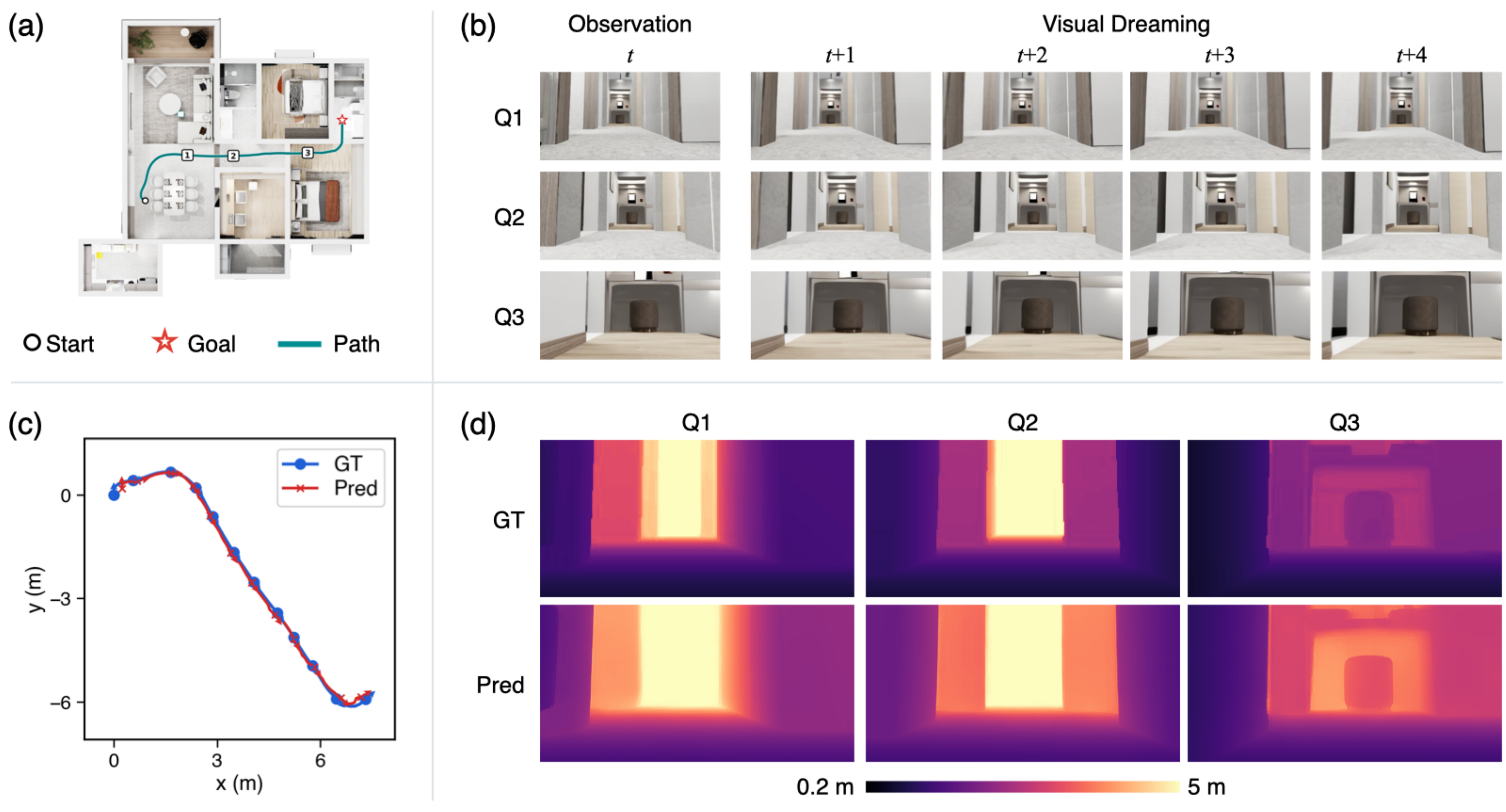}
\caption{\textbf{Navigation result in Home scene 16, episode 94.}}
\label{fig:app_vis_3}
\end{figure}

\begin{figure}[p]
\centering
\includegraphics[width=\linewidth]{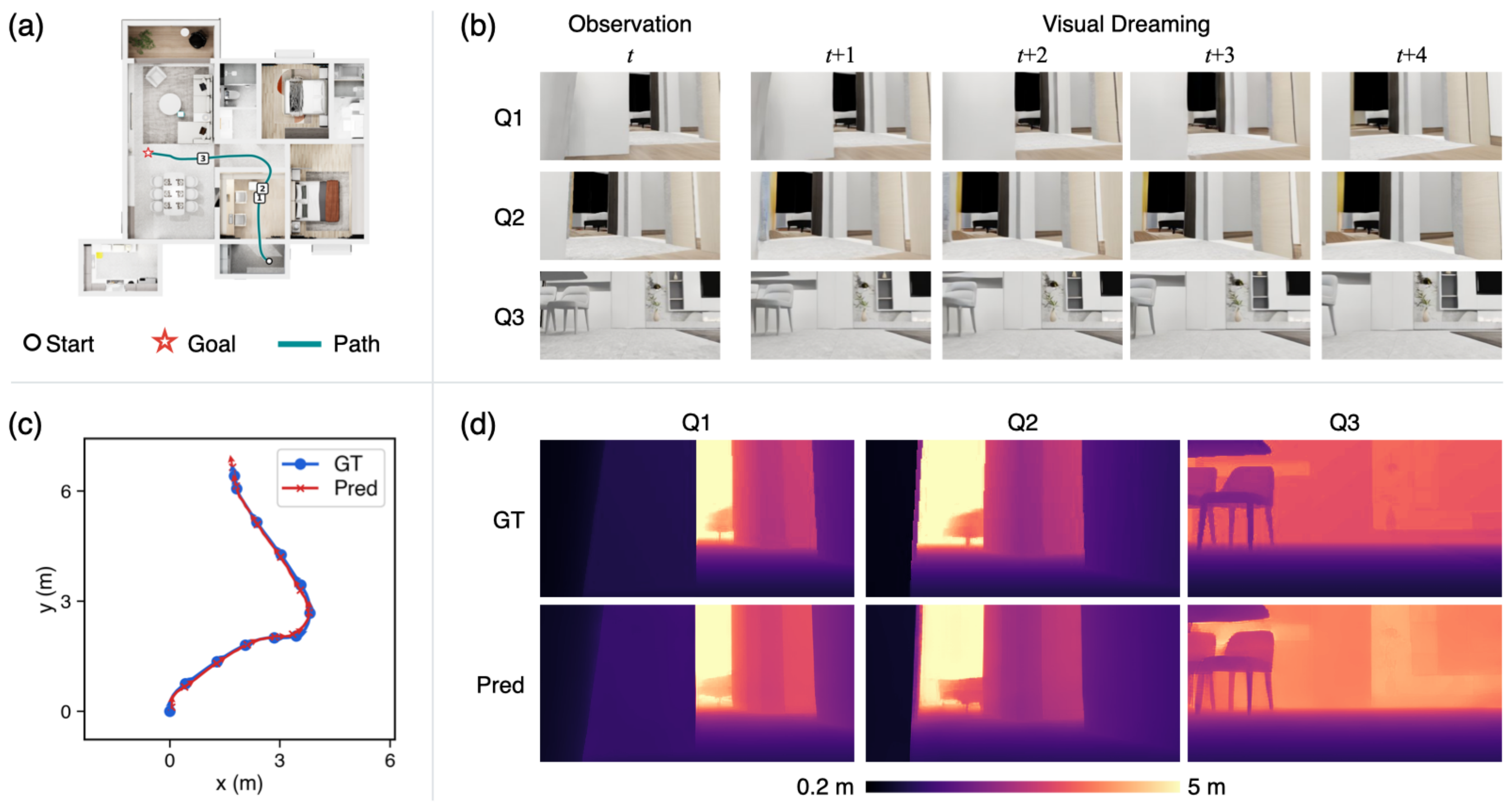}
\caption{\textbf{Navigation result in Home scene 16, episode 96.}}
\label{fig:app_vis_4}
\end{figure}

Figures~\ref{fig:app_vis_1}--\ref{fig:app_vis_4} show additional closed-loop navigation results in simulation, with panels as in Figure~\ref{fig:navigation_visualization}. The four episodes are taken from two Home scenes of the InternVLA-N1 start--goal benchmark.

\end{document}